\documentclass{article}
\usepackage{spconf,amsmath,graphicx}
\usepackage{amssymb}

\usepackage{enumitem}
\setlist{nosep, leftmargin=14pt}

\usepackage{mwe} 

\title{A Hierarchical Ensemble Inference Pipeline for Robust White Blood Cell Classification Under Domain Shifts}
\name{Ruyi Dai\textsuperscript{\textup{1},\dag}, Tingkwong Ng\textsuperscript{\textup{1},\dag}, Hao Chen\textsuperscript{\textup{1,2,3,4,5},*}\thanks{\dag Co-first-authors; *Corresponding author.}}
\address{
    \textsuperscript{1}Department of Computer Science and Engineering, \\
    \hspace{1em} The Hong Kong University of Science and Technology, Hong Kong, China \\
    \textsuperscript{2}Department of Chemical and Biological Engineering, \\
    \hspace{1em} The Hong Kong University of Science and Technology, Hong Kong, China \\
    \textsuperscript{3}Division of Life Science, Hong Kong University of Science and Technology, Hong Kong, China \\
    \textsuperscript{4}HKUST Shenzhen-Hong Kong Collaborative Innovation Research Institute, Futian, Shenzhen, China \\
    \textsuperscript{5}State Key Laboratory of Nervous System Disorders, \\
    \hspace{1em} The Hong Kong University of Science and Technology, Hong Kong, China
}
\begin{document}
%

\maketitle
\begin{abstract}
Automated white blood cell (WBC) classification is essential for scalable leukaemia screening. However, real-world deployment is challenged by domain shifts caused by staining protocols, scanner characteristics, and inter-laboratory variability, which often degrade model performance. The White Blood Cell Classification Challenge (WBCBench) at ISBI 2026 aims to advance robust WBC recognition, with a focus on accurately identifying blast cells and other clinically critical rare subtypes. We propose a memory-augmented, hierarchical ensemble pipeline for WBC classification under domain shifts, leveraging a feature bank and a DinoBloom backbone fine-tuned with LoRA. Our three-stage inference hierarchy combines k-nearest neighbors (kNN) retrieval at each level, reducing over-reliance on any single decision. Evaluated on the WBCBench dataset, our method ranks within the top ten by macro F1-score in the final testing phase.
\end{abstract}
\section{Introduction}
\label{sec:intro}
Leukaemia is a life-threatening hematological malignancy, accounting for approximately 4\% of cancer-related deaths and ranking among the leading causes of cancer mortality in both males and females~\cite{siegel2022cancer}. The World Health Organization statistics reported 2.9 new cases and 1.2 deaths per 100,000 people in individuals aged 0 to 24 years in 2022. Peripheral blood smear morphology analysis is crucial for diagnosis and treatment evaluation, but manual microscopic assessment is time-consuming, labor-intensive, and subject to inter-observer variability. 

Deep learning models offer potential for streamlining screening and improving diagnostic consistency~\cite{acevedo2019recognition,asghar2023automatic}. However, CNN-based approaches have shown limited gains in abnormal WBC detection accuracy~\cite{tayebi2021histogram,ammar2022feature,xing2023artificial}. While Vision Transformers (ViTs) improve global feature capture, existing evaluations remain largely confined to single-dataset splits with inadequate validation under domain shifts and patient-level generalization~\cite{katar2023explainable,rubin2023transforming}. The ISBI 2026 WBCBench Challenge~\cite{wbcbench2026} targets clinically realistic domain shifts and class imbalance, emphasizing robust cross-domain WBC classification with reliable performance on rare subtypes.

To address these challenges, we introduce a hierarchical ensemble framework that leverages LoRA-adapted DinoBloom~\cite{koch2024dinobloom} embeddings for coarse-to-fine inference. Our approach combines retrieval-based kNN prediction within the hierarchy, and aggregates outputs across multiple splits via majority voting to improve robustness under domain shifts. The main contributions are summarized as follows:
\begin{enumerate}[label=\arabic*., leftmargin=*, itemsep=0pt, topsep=0pt, parsep=0pt]
  \item We propose a coarse-to-fine hierarchical ensemble inference strategy for robust WBC classification under domain shifts.
  \item  We fine-tune a DinoBloom backbone
  with LoRA to obtain transferable embeddings that support retrieval-based inference under domain shifts.
 
  \item We further employ multi-split majority voting to stabilize predictions and improve performance on long-tailed classes.

\end{enumerate}

\section{METHODOLOGY}
\begin{figure*}[htb]
  \centering
  \includegraphics[width=0.73\textwidth]{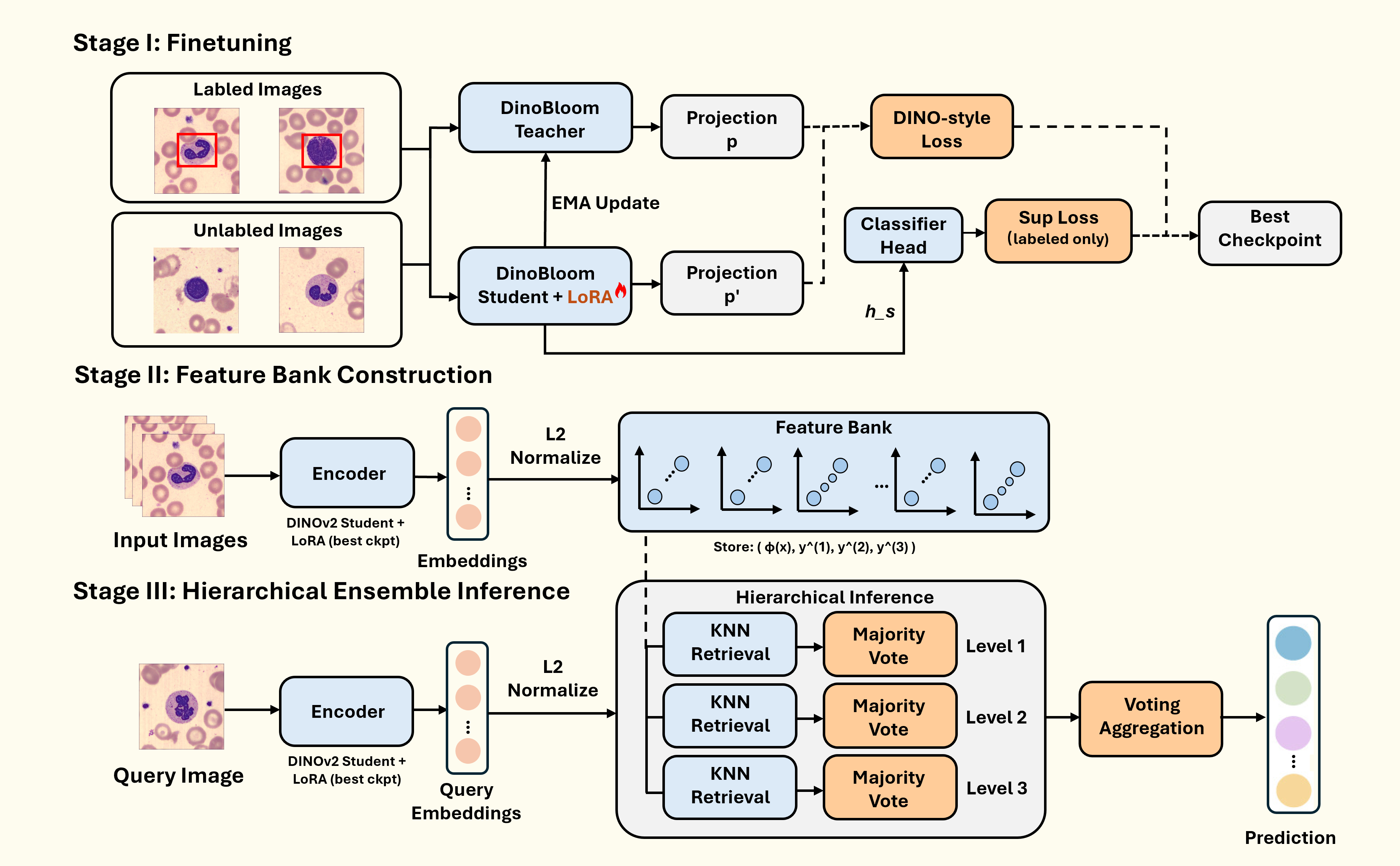}
  \caption{Overview of the proposed three-stage pipeline.}
  \label{fig1}
\end{figure*}

Fig.~\ref{fig1} provides an overview of our three-stage pipeline designed to handle class imbalance and domain shift in white blood cell classification.
\vfill
\pagebreak

\subsection{Dataset Preparation}
WBCBench Challenge~\cite{wbcbench2026} provides labeled WBC images for a 13‑class task. To mitigate class imbalance, we use a three‑level hematology‑aligned hierarchy and train with both leaf and parent labels. We merge the development splits and run 5‑fold cross‑validation for training and selection. PBC~\cite{acevedo2020dataset} and Raabin‑WBC~\cite{kouzehkanan2021raabin} are added as auxiliary data after label mapping to this taxonomy. The official test split is kept held out for final evaluation.

Level‑1 lineages are Myeloid, Lymphoid, and Blast. Myeloid leaves: PMY, MY, MMY, BNE, SNE, MO, EO, BA. Lymphoid leaves: PLY, LY, PC, VLY. Blast leaf: BL. Images are center‑cropped to the cell region and resized to the input resolution. Training uses a two‑view teacher–student setup with independent geometric and appearance augmentations. Validation and testing use deterministic preprocessing for reproducible evaluation.

\subsection{Method Overview}

\subsubsection{Stage I: Fine-tuning}
We initialize the DinoBloom backbone with its pretrained checkpoint and adapt it via LoRA~\cite{hu2022lora} fine-tuning. Let \(f_{\theta}(\cdot)\) denote the backbone that maps an input image \(x\) to a global embedding \(z\in\mathbb{R}^{d}\).
We adopt an EMA teacher--student scheme with two independently augmented views \(x^{t}\) and \(x^{s}\):
\begin{equation}
z^{t}=f_{\theta_{t}}(x^{t}),\qquad
z^{s}=f_{\theta_{s}}(x^{s}),
\label{eq:ts-embed}
\end{equation}
where the teacher shares the same architecture and is updated as the EMA of the student,
\begin{equation}
\theta_{t}\leftarrow m\,\theta_{t}+(1-m)\,\theta_{s}.
\label{eq:ema}
\end{equation}
with EMA momentum \(m\in(0,1)\).
We optimize the LoRA parameters together with lightweight projection and classification heads.

We enforce teacher--student alignment using a DINO-style loss,
\begin{equation}
\mathcal{L}_{\text{dino}}
= -\frac{1}{B}\sum_{i=1}^{B}\sum_{k=1}^{K}
p^{t}_{ik}\log p^{s}_{ik},
\label{eq:dino}
\end{equation}
where \(K\) is the projection-head output dimension, and \(p^{t}\) and \(p^{s}\) are teacher and student output distributions obtained by applying temperature-scaled softmax to the projection-head outputs.
For labeled samples \(y\in\{1,\dots,C\}\) with \(C=13\), we train a classifier head \(h(\cdot)\) using a class-balanced cross-entropy:
\begin{equation}
\mathcal{L}_{\text{sup}}
= -\frac{1}{B_{\ell}}\sum_{i\in\mathcal{I}_{\ell}} w_{y_i}\,
\log \frac{\exp \left(h(z^{s}_{i})_{y_i}\right)}
{\sum_{c=1}^{C}\exp \left(h(z^{s}_{i})_{c}\right)}.
\label{eq:sup}
\end{equation}

The final objective is
\begin{equation}
\mathcal{L}
= \lambda_{\text{dino}}\mathcal{L}_{\text{dino}}
+ \lambda_{\text{sup}}\mathcal{L}_{\text{sup}}.
\label{eq:total}
\end{equation}
where \(\lambda_{\text{dino}}\) and \(\lambda_{\text{sup}}\) balance self-supervised alignment and supervised classification.
We retain the checkpoint that maximizes the macro F1-score on the evaluation split:
\begin{equation}
\text{Macro F1-score}=\frac{1}{C}\sum_{c=1}^{C}\text{F1}_{c}.
\label{eq:macrof1}
\end{equation}

\subsubsection{Stage II: Feature Bank Construction}
Using the selected checkpoint, we extract \(\ell_{2}\)-normalized embeddings and construct a feature bank,
\begin{equation}
\phi(x)=\frac{f_{\theta_{s}}(x)}{\|f_{\theta_{s}}(x)\|_{2}}.
\label{eq:featbank}
\end{equation}

For each database image $x_j$, we store its normalized embedding $\phi(x_j)$ and the corresponding hierarchical labels $y_j^{(1)}$, $y_j^{(2)}$, and $y_j^{(3)}$ in the feature bank.
\subsubsection{Stage III: Hierarchical kNN Inference}
Given a query image \(x\), we compute \(q=\phi(x)\) and retrieve its \(k\)-nearest neighbors \(\mathcal{N}_{k}(q)\) from the feature bank using cosine similarity. Each neighbor \(j\in\mathcal{N}_{k}(q)\) is associated with a three-level hierarchical label \(\big(y^{(1)}_{j},y^{(2)}_{j},y^{(3)}_{j}\big)\), where level \(l=3\) corresponds to the leaf classes.

Inference proceeds from coarse to fine. For \(l=1\), we predict by majority voting:
\begin{equation}
\hat{y}^{(1)}=\operatorname*{mode}\left\{y^{(1)}_{j}\right\}_{j\in\mathcal{N}_{k}(q)}.
\label{eq:knn-level1}
\end{equation}
For finer levels \(l\in\{2,3\}\), we enforce hierarchical consistency by restricting candidates to the children of the predicted parent. Let \(\mathrm{Ch}^{(l)}(\cdot)\) denote the valid children at level \(l\) given a parent prediction at level \(l-1\). We vote only over neighbors whose level-\(l\) labels fall in \(\mathrm{Ch}^{(l)}\!\left(\hat{y}^{(l-1)}\right)\):
\begin{equation}
\hat{y}^{(l)}=\operatorname*{mode}\left\{y^{(l)}_{j}\right\}_{j\in\mathcal{N}_{k}(q),\; y^{(l)}_{j}\in \mathrm{Ch}^{(l)}(\hat{y}^{(l-1)})}.
\label{eq:knn-level}
\end{equation}
We repeat this procedure until the leaf level and output \(\hat{y}=\hat{y}^{(3)}\).

\section{EXPERIMENTS AND  RESULTS}
\subsection{Implementation}
Our method is implemented in PyTorch. We fine-tune the DinoBloom weights using LoRA on a single NVIDIA A100 for 100 epochs with a batch size of 16. We optimize the model with AdamW using a learning rate of $1\times10^{-5}$, a weight decay of $1\times10^{-2}$, and an EMA momentum of 0.999. Inference is performed on an NVIDIA RTX 6000 Ada using hierarchical kNN voting.
\begin{table}[t]
\centering
\caption{Quantitative evaluation on WBCBench under different settings. Best result is \textbf{bold}; second-best is \underline{underlined}.}
\label{tab1}
\renewcommand{\arraystretch}{1.15}
\setlength{\tabcolsep}{4pt}

\begin{tabular}{l c c c}
\hline
\textbf{Method} & \textbf{Pipeline} & \textbf{Finetuning} & \textbf{MF1} \\
\hline
ResNet-50  & supervised                     & full & 0.635 \\
Swin-T     & supervised                     & full & 0.643 \\
ViT-B      & supervised                     & full & 0.631 \\
ConvNeXt-L & hybrid (SSL+sup)               & full & 0.679 \\
ConvNeXt-L\textsuperscript{1} & hybrid (SSL+sup) & LoRA & 0.676 \\
ConvNeXt-L\textsuperscript{2} & hybrid (SSL+sup) & LoRA & 0.677 \\
ConvNeXt-L\textsuperscript{3} & hybrid (SSL+sup) & LoRA & 0.678 \\
DinoBloom\textsuperscript{1}  & hybrid (SSL+sup) & LoRA & \underline{0.681} \\
DinoBloom\textsuperscript{2}  & hybrid (SSL+sup) & LoRA & \textbf{0.682} \\
DinoBloom\textsuperscript{3}  & hybrid (SSL+sup) & LoRA & 0.680 \\
\hline

\end{tabular}

\vspace{2pt}
{\footnotesize \(^{1,2,3}\): different train/eval splits and augmentation settings}
\end{table}

\subsection{Quantitative Results}
We evaluate all methods using the macro F1-score (MF1). Table~\ref{tab1} summarizes the performance of different backbones and training pipelines. Among the supervised baselines, Swin-T achieves an MF1 of 0.643, outperforming ResNet-50 at 0.635 and ViT-B at 0.631.

We develop three in-house methods for comparison, including DinoBloom training with LoRA fine-tuning, ConvNeXt hybrid training with full fine-tuning, and a two-stage ConvNeXt pipeline with self-supervised pre-training followed by supervised LoRA fine-tuning. ConvNeXt-L with full fine-tuning achieves an MF1 of 0.679. We also evaluate ConvNeXt-L with LoRA fine-tuning under different evaluation-split augmentation settings, where ConvNeXt-L\textsuperscript{1}, ConvNeXt-L\textsuperscript{2}, and ConvNeXt-L\textsuperscript{3} obtain MF1 scores of 0.676, 0.677, and 0.678, respectively. DinoBloom with LoRA fine-tuning delivers the strongest performance, where DinoBloom\textsuperscript{2} achieves the overall MF1 of 0.682. The results show that DinoBloom with LoRA achieves the best MF1, slightly outperforming full fine-tuned ConvNeXt-L while being more parameter efficient.

\begin{table}[t]
\centering
\caption{Ablation on hierarchical inference and the number of ensemble models.}
\label{tab2}
\renewcommand{\arraystretch}{1.15}
\setlength{\tabcolsep}{4pt}

\begin{tabular}{c c c}
\hline
\textbf{\#Ensemble model} & \textbf{W/O Hierarchy} & \textbf{W/ Hierarchy} \\
\hline
1 & 0.596 & 0.676 \\
2 & 0.608 & 0.677 \\
3 & 0.614 & 0.678 \\
4 & 0.617 & 0.679 \\
5 & 0.621 & 0.681 \\
6 & 0.623 & 0.681 \\
7 & 0.625 & \textbf{0.682} \\
\hline
\end{tabular}
\end{table}

\subsection{Ablation Study}
Table~\ref{tab2} further investigates the effects of split ensembling and hierarchical inference. This ensemble aggregates predictions from our three in-house methods under different split-dependent settings. Without hierarchical constraints, MF1 increases gradually from 0.596 with one split to 0.625 with seven splits, indicating that split ensembling provides a modest but consistent robustness gain. In contrast, enabling hierarchical inference yields substantially higher MF1 across all split numbers, increasing from 0.676 with one split to 0.682 with seven splits. This highlights the effectiveness of hierarchical inference under class imbalance, where coarse-to-fine constraints reduce cross-level errors that disproportionately affect rare classes and help mitigate majority-class bias, leading to more robust and accurate predictions.

\section{Conclusion}
In this work, we propose a hierarchical kNN-based inference framework for fine-grained white blood cell classification on WBCBench. This task is challenging due to subtle inter-class differences and limited labeled data, which can make standard end-to-end classifiers less robust. Our approach fine-tunes a pretrained backbone to learn transferable representations, then performs inference by constructing a feature bank and aggregating split-ensemble predictions via hierarchical voting, improving robustness with minimal overhead. Our final model achieves a macro-F1 score of 0.682 on the competition test set. Ablation studies show that both hierarchical inference and split ensembling contribute to the performance gains, with more splits providing consistent improvements. Future work will explore building a foundation model for blood cell analysis, enabling more generalizable representations across cell types, tasks, and domains.

\bibliographystyle{IEEEbib}
\bibliography{refs}

\end{document}